\documentclass[10pt, a4paper]{article}
\usepackage[utf8]{inputenc}
\usepackage[dvipsnames]{xcolor}

\usepackage[shortlabels]{enumitem}

\usepackage{amsmath}
\usepackage{bbm}

\newcommand{\publicmeetings}{\texttt{\small public\_meetings}}

\usepackage{adjustbox}
\usepackage{multirow}
\usepackage{booktabs}

\newcommand{\submidrule}{\midrule[0.0em]}
\usepackage{fourier} 
\usepackage{array}
\usepackage{makecell}
\usepackage{float}

\newcommand{\subhead}[2]{\thead{\textbf{#1} \\ #2}}
\DeclareMathOperator{\score}{\mathit{score}}

\DeclareMathOperator{\T}{\mathcal{T}}
\DeclareMathOperator{\R}{\mathcal{R}}
\DeclareMathOperator{\lenT}{\mathit{I}}
\DeclareMathOperator{\lenR}{\mathit{J}}

\DeclareMathOperator{\TS}{\hat{\mathcal{T}}}
\DeclareMathOperator{\RS}{\hat{\mathcal{R}}}
\DeclareMathOperator{\lenTS}{\mathit{M}}
\DeclareMathOperator{\lenRS}{\mathit{N}}

\DeclareMathOperator{\tseg}{\mathit{T}}
\DeclareMathOperator{\rseg}{\mathit{R}}
\DeclareMathOperator{\tsen}{\mathit{t}}
\DeclareMathOperator{\rsen}{\mathit{r}}
\DeclareMathOperator{\alignmentfct}{\mathit{alignment}} % alignment function

\newcommand{\up}[1]{\textsuperscript{#1}}

\makeatletter
\DeclareFontEncoding{LS1}{}{}
\DeclareFontEncoding{LS2}{}{\noaccents@}
\DeclareFontSubstitution{LS1}{stix}{m}{n}
\DeclareFontSubstitution{LS2}{stix}{m}{n}
\makeatother

\DeclareMathAlphabet\mathscr{LS1}{stixscr}{m}{n}
\DeclareMathAlphabet\mathcal{LS2}{stixcal}{m}{n}
\DeclareMathOperator*{\argmax}{arg\,max}

\usepackage{lrec}
\usepackage{graphicx}
\usepackage{tabularx}
\usepackage{soul}
\usepackage{epstopdf}

\usepackage{hyperref}
\usepackage{xstring}

\usepackage{tikz}
\usetikzlibrary{arrows}
\usetikzlibrary{arrows.meta}
\usepackage{subfig}

\title{Align then Summarize: Automatic Alignment Methods for Summarization Corpus Creation}

\name{Paul Tardy\up{1, 2}\quad David Janiszek \up{1, 3}\quad Yannick Estève \up{4}\quad Vincent Nguyen \up{2}}

\address{
    \up{1} LIUM -- Le Mans Université \\ 
    \up{2} Ubiqus Labs\\
    \up{3} Université Paris Descartes \\ 
    \up{4} LIA -- Avignon Université \\
      pltrdy@gmail.com, 
      david.janiszek@parisdescartes.fr,\\
      yannick.esteve@univ-avignon.fr,
      vnguyen@ubiqus.com\\
}
\abstract{
Summarizing texts is not a straightforward task.
Before even considering text summarization, one should determine what kind of summary is expected. How much should the information be compressed? Is it relevant to reformulate or should the summary stick to the original phrasing? State-of-the-art on automatic text summarization mostly revolves around news articles. We suggest that considering a wider variety of tasks would lead to an improvement in the field, in terms of generalization and robustness. 
We explore meeting summarization: generating reports from automatic transcriptions. Our work consists in segmenting and aligning transcriptions with respect to reports, to get a suitable dataset for neural summarization.
Using a bootstrapping approach, we provide pre-alignments that are corrected by human annotators, making a validation set against which we evaluate automatic models. 
This consistently reduces annotators' efforts by providing iteratively better pre-alignment and maximizes the corpus size by using annotations from our automatic alignment models.
Evaluation is conducted on \publicmeetings, a novel corpus of aligned public meetings. We report automatic alignment and summarization performances on this corpus and show that automatic alignment is relevant for data annotation since it leads to large improvement of almost +4 on all ROUGE scores on the summarization task. 
\newline }

\begin{document}
\maketitleabstract

\section{Introduction}

% DEFINING SUMMARIZATION
Automatic Text Summarization is the task of producing a short text that captures the most salient points of a longer one. However, a large variety of tasks could fit this definition. 

% % SUMMARIZATION IS A VERY WIDE TASK -- all corpora are different
% First, we often consider two approches: \\
% (i) \textit{extractive summarization}, that select parts of the source as the summary i.e. a kind of cut and paste process; \\
% (ii) \textit{abstractive summarization} that generates a novel text i.e. read then write.

%% Yannick: ne pas appuyer sur la différence ext/abs
% The abstractive approach tends to be closer to the human process, hence became an active field of  research. However, they are not mutually exclusive, and the abstractive process benefits from \textit{extractive} mechanisms (which we can seen as ``read, highlight then write``). Recent research is heading in that direction, in particular copy-mechanisms that decides whether to generate a word or copy it from the source \cite{Vinyals2015,Nallapati2016,see2017}, and extractive pre-processing \cite{Liu2018,Gehrmann2018}. \textit{Abstractiveness} is often reported in term of percentage of target present in the source.

Many factors are critical in the summarization process, such as whether to rephrase the source (abstractiveness) or use part of the source as-is (extractiveness);  the length ratio of target and source (compression factor); the source and target lengths and their variances; and the information distribution -- i.e. how important information is distributed along the text. 
% For example, the CNN/DailyMail corpus has quite long sources (up to 800 words) that leads to short summaries (around 100 words, high compression factor), without much rephrasing (low abstractiveness). In the other hands, when summarizing meetings, we may work with both short, and very long interventions (high variance, medium mean), that should be rewriting, from the oral to the written form, leading to a high abstractiveness but a quite low compression factor, depending on how exhaustive the report is. Also, news texts tends to concentrate their most informative content in the first sentences, which is why research on CNN/DailyMail tends to truncate the source to 400 words (instead of up to 800) without loss of performance. On the other hand, this assumption does not hold for meeting data, we have to consider the whole input to not miss important information.

Most of summarization benchmarks \cite{see2017,Paulus2017,Gehrmann2018} rely on news articles from CNN and DailyMail \cite{Hermann2015a,Nallapati2016} which exhibit particular characteristics such as: (i) being quite extractive i.e. picking portions of text from the source, the opposite of abstractive \cite{Liu2018}; (ii) a high compression factor with the summary being up to 10 times shorter than the source \cite{Liu2018}; (iii) a low variance in both source and target length, and (iv) concentrating information mostly at the beginning of the article: for example, papers working on the CNN-DailyMail corpus \cite{Hermann2015a,Nallapati2016} often truncate the article to the first 400 words of the article \cite{see2017,Gehrmann2018,ziegler2019}, ignoring up to half of it.

In contrast, we explore meeting data, using transcription as the source, and the meeting report as the target. 

Contrary to news articles, there is high variance in the length of speaker interventions; the data need to be rephrased into a written form (thus, an abstractive process by nature), and to be informative throughout.
In this work, we focus on so-called \textit{exhaustive reports}, which are meant to capture all the information and keep track of speaker interventions. Information itself is not summarized, but the speech is compressed from an oral form to a written one. Thus, the compression factor is lower than in news tasks but variance remains high, depending on how verbose the intervention is.

The data at hand consist of (i) exhaustive reports produced by Ubiqus in-house editors, (ii) full audio recording of the meeting. An automated transcript is produced from the latter with an automatic speech recognition system very close to the one described \cite{Hernandez2018,Meignier2010} but trained for French language from internal data.

Such data are not suitable for summarization learning as-is, therefore we propose to segment it at the intervention level (\textsl{i.e}. what is said from one speaker until another one starts). It is particularly convenient since the nature of our dataset ensures that all interventions remain (apart from very short ones) and chronological order is preserved in both the transcription and the report. Reports explicitly mention speakers, making segmentation trivial and error-free for that part. Transcriptions do not have such features so we present an alignment process that maps interventions from the reports with its related transcription sentences based on similarity.

We bootstrap the corpus creation, iterating between automatic pre-alignment generations and corrections from human annotators. We aim at jointly minimizing human effort while fine-tuning automatic alignment models to eventually use alignment models for automatic data annotation.

In this paper, we present a methodology for building a summarization corpus based on the segmentation and alignment of reports and transcriptions from meetings using a bootstrapping approach. 
We also present \publicmeetings, a novel public meeting dataset, against which we evaluate both automatic alignment and summarization. 
Summarization models are first trained on the gold set -- from human annotator --, and then using automatic annotations with our automatic alignment models which outperform the baseline by a large margin (almost +4 on all considered ROUGE metrics). Source code, data and reproduction instructions can be found at:\\
\url{https://github.com/pltrdy/autoalign}.

\section{Related Work}

% \todo{Long story short, I found similar works, but too different to be really comparable. In particular, if the starting hypothesis are not respected i.e. summary by intervention, chronology etc.
% However, TextTiling \cite{Hearst1997} provide an unsupervised segmentation method that we can evaluate. However, it only operates on a single text, thus, it's disavantaged in the comparison.}

This work aims to jointly segment two related files -- a transcription and a report of the same meeting -- so that the $i$-th segment of the report actually corresponds to the $j$-th segment of the transcription. 

% \cite{Hearst1997}: tl;dr: segmentation algorithm for single document; comparing adjacent block of text to find borders (aka. subtopic shifts). `` multi-paragraph units that represent passages, or subtopics. The discourse cues for identifying major subtopic shifts are patterns of lexical co-occurrence and distribution. ``

Since report side segmentation is simple thanks to its structure, we focus on the transcription side. Bearing that in mind, the task is similar to a linear segmentation problem, i.e. finding borders between segments. \cite{Hearst1997} proposed \textsc{TextTiling}, a linear segmentation algorithm that compares adjacent blocks of text in order to find subtopic shifts (borders between segments) using a moving window over the text and identifying borders by thresholding. 
% \cite{Choi}: tl;dr: linear segmentation aka. C99
\textsc{C99}, as proposed by \cite{Choi}, uses similarity and ranking matrices instead, then clustering to locate topic boundaries. 

% NOTE: not relevant here.
% \cite{Pevzner}: tl;dr: discuss metrics in segmentation tasks; other that precision, recall or so-called metric $P_k$ they introduce \texttt{WindowDiff} that does not penalize false-negative as hard (quite out topic, just that I considered WindowDiff at some point)

% NOTE: good review, not mandatory here
% \cite{Sitbon2004} tl;dr: sota review of the time for linear segmentation methods. 

% \cite{Banerjeea} tl;dr: extending TextTiling (how?) for Topic Boundary Detection in Meetings. Based on audio signal. Weak on atypical structure e.g. long discourse and short question, which we meet.
% \cite{Song2016} tl;dr: TextTiling w/ word embeddings for dialogue segmentation w.r.t a query.

\textsc{TextTiling} has been extended (i) to audio signals \cite{Banerjeea} but is said to lack robustness to atypical participant behavior (which is common in our context); (ii) to work with word embeddings in order to capture similarity between query and answer in a dialogue context \cite{Song2016}. \cite{Alemi2015} also explore word embedding use in segmentation by incorporating it into existing algorithms and showing improvements. \cite{Badjatiya2018} address the segmentation task with an end-to-end attention-based neural approach. While such an approach could be investigated in the future, we could not consider it in this work due to the lack of reference data.

\cite{Glavas} use semantic relatedness graph representation of text then derive semantically coherent segments from maximal cliques of the graph. One issue of this approach is that searching for large segments in big texts requires decreasing the threshold which exponentially increases computational cost, eventually making our task intractable.

\subsection{Alignment}
% \cite{Barzilay}: tl;dr: propose a method for so called ``Sentence Alignment for Monolingual Comparable Corpora``. While the idea sounds similar, it focuses on selecting aligned sentences in two texts, i.e. extracting sentence pairs from texts that share the same meaning. This, could be useful in order to build a sentence summarization corpora, but we, want to align the whole texts instead of extracting exactly aligned pairs.
% \cite{Nelken2006}: tl;dr: similar to \cite{Barzilay}, better tho.

Alignment has already been studied for corpus creation. In particular, \cite{Barzilay,Nelken2006} extract related segments from the \textit{Encyclopedia Britannica} and \textit{Britannica Elementary} (a simpler version). It is different from our work since we are looking for a total alignment, i.e. both documents must be fully aligned, not just partially extracted.

% \notes{"Logical" alignment, ie NLI, predicates,  }
% \cite{MacCartney2008}: \todo{alignment for NLI, not really related}
% \cite{Roth2012}: tl;dr: word alignment
% \cite{Yao}: tl;dr: word, related to \cite{MacCartney2008} and \cite{Roth2012}

Furthermore, alignment of oral speech to its written form has been studied by \cite{Braunschweiler2010} in the context of audio books and by \cite{Lecouteux2011} for subtitles and transcripts (e.g. of news report) in order to improve Automatic Speech Recognition engines. While such approaches sound similar to ours, they mostly look for exact matches rather than an approximate alignment of asymmetrical data, based on textual similarity. 
% \todo{tl;dr: aligning audio (audio books) with text description, it's a perfect match rather than a semantic similarity}

% \cite{Lecouteux2011}: \todo{Leverage the amount of close to perfect match data such a subtitles and trascripts. In this case, the actual speech, if not always exactly match the transcription remains really close. The alignment helps ASR to locate transcription mismatchs (aka. transcription island) and correct transcription}

% \subsection{Other}
% \todo{}
% \begin{enumerate}
%     \item \cite{Matuschek} tl;dr: text similarity w/ dynamic time warping, similar in the idea to our model (i.e. using dynamic programming to align scores). Also, DTW in itself is used as a metric. 
% \end{enumerate}

\subsection{Summarization Datasets}
% \begin{table*}[!ht]
% \begin{center}
% \begin{tabular}{ccccc} % c|}
% \toprule
%     Dataset
%     & Input
%     & Output
%     & \# Examples
%     & ROUGE-1 R
% \\ \midrule
%     CNN/DailyMail \cite{Nallapati2016} 
%         & $10^2$ -- $10^3$
%         & $10^1$
%         & $10^5$
%         & $76.1$
% \\  WikiSum \cite{Liu2018} 
%         & $10^2$ -- $10^6$
%         & $10^1$ -- $10^3$
%         & $10^6$
%         & $59.2$
% \\ \midrule 
%     Gold \textit{filtered}(ours)
%         & $261$
%         & $198$
%         & $2.10^3$ 
%         & $54.0$
% \\\bottomrule
% \end{tabular}
% \caption{Scores on the \publicmeetings  test set of automatic summarizations models trained on human references only vs. data augmentation with automatic alignment}
% \label{table:results_summarization_publicmeetings}
%  \end{center}
% \end{table*}

\cite{Hermann2015a,Nallapati2016} proposed the first multi-sentence summarization dataset, with more than 280.000 training pairs. Sources are up to 800 words long (but are often truncated to the first 400 words \cite{see2017,Gehrmann2018,ziegler2019}) and the target is around 50 words on average. A similar dataset based on NY Times articles was presented by \cite{Paulus2017}, with three times more training pairs, sources of 800 words and targets of 45 words on average. \cite{Liu2018} work on generating Wikipedia introductions (known as leads) from reference articles and web-crawled data. Both inputs and outputs are several orders of magnitude longer: sources can be up to $10^6$ words and targets are in the $10^1$ -- $10^3$ range.

In our context, we are dealing with limited resources, in particular with respect to ready to train data -- which motivated this paper. Our dataset comprises 20,000 gold standard training pairs, and up to 60,000 pairs when taking into account all the automatically aligned data. We currently filter training pairs in order to contain fewer than 1000 words and 50 sentences. Future work would explore a wider range of segment lengths.
\section{Methods}
Our task consists in finding the best alignment between a meeting transcription $\T=\left\{\tsen_1,\ldots, \tsen_{\lenT}\right\}$ and the related human written report $\R=\left\{\rsen_1,\ldots, \rsen_{\lenR}\right\}$.

Both documents are \textit{segmented} into mutually exclusive sets of sentences
$\TS = \left\{\tseg_1, \ldots, \tseg_{\lenTS} \right\}$, $\RS = \left\{\rseg_1, \ldots, \rseg_{\lenRS}\right\}$.

\textit{Alignment} maps each transcription segment $\tseg_m\in \TS$ to exactly one report segment $\rseg_n \in \RS$ based on sentence-level similarities $S_{i, j} = \score(\tsen_i, \rsen_j)$, with $\tsen_i \in \T$, $\rsen_j \in \R$.

The alignment process is a pipeline of different modules. The first one reads the data; the second one independently segments each side -- respectively report and transcription; the third one computes similarity scores in order to find the alignment that maximizes the overall score. This section presents those modules.

\subsection{Segmentation}
Segmentation consists in finding borders in texts such that each segment can be processed independently. The segmentation granularity should be fine enough for segments to be not too long (which would make learning more difficult, and result in fewer training pairs) and coarse enough to remain relevant (very short segments cannot be meaningfully summarized).
We consider speaker interventions -- i.e. uninterrupted speech of only one speaker -- to be an appropriate segmentation level. In particular, we make the assumption that the task of writing a report can roughly be divided into sub-tasks consisting of reports for each intervention, which is a close approximation of \textit{exhaustive reports}.

On the report side, each speaker's intervention may be explicitly mentioned using special tags in the document (one particular style applied to names), or identified with rule-based identification e.g. looking for `Mr.`, `Ms.` etc.

On the transcription side, segments are groups of sentences defined by the automatic speech recognition system. 

% \notes{
%     \begin{enumerate}
%         \item SRC: segmenting CTM based on "<start" tags. 
%         \item TGT: segmenting docx based on tags (paragraphs, titles, and presence of style "nom")
%     \end{enumerate}
% }
\subsection{Text Representation and Similarity function}
The alignment process consists in finding, for each transcription segments $\tseg_m$, its related report segment $\rseg_n$, in other words the function:
\begin{align*}
    \alignmentfct(m) &= n\text{, }\forall m \in [1, \lenTS]\text{, }n \in [1, \lenRS]
\\  \alignmentfct(m) &\leq \alignmentfct(m+1)
\end{align*}

We consider a sentence-level similarity matrix $S$ between the transcription $\T$ and the report $\R$ such as $S_{i,j} = score(\tsen_i, \rsen_j)$ with $(\tsen_i, \rsen_j)\in \T \times \R$.

\begin{figure}[h]
\begin{center}
%\fbox{\parbox{6cm}{
%This is a figure with a caption.}}
% \includegraphics[scale=0.5]{fig/postedit_similarity.png} 
\includegraphics[width=\linewidth]{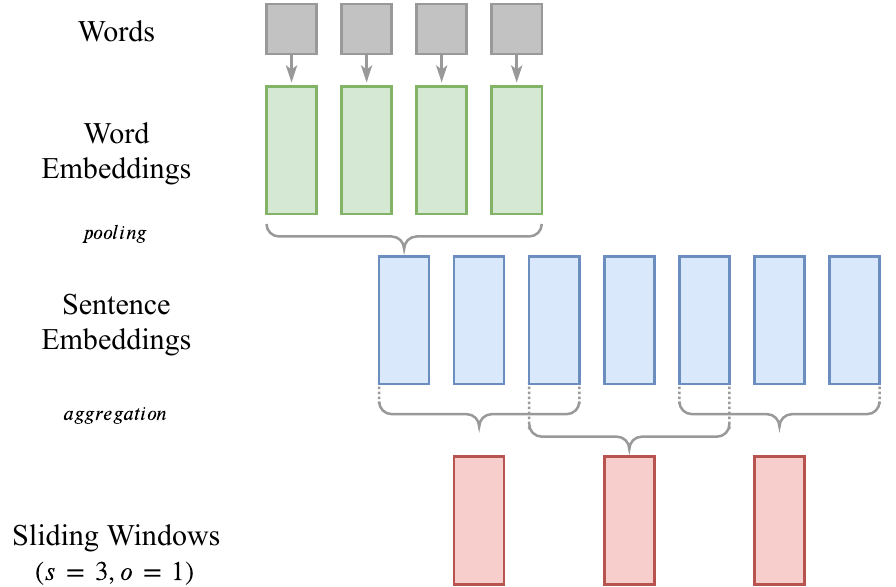} 
\caption{Text Representations: from words to sliding-windows.}
\label{fig:methods_representations}
\end{center}
\end{figure}

For the $\score$ function, we experimented with (i) $ROUGE$ from \cite{Lin2004}; (ii) cosine similarity on $tf\cdot idf$ representations ; (iii) cosine similarity based on word embedding vectors. A $pooling$ function (typically a sum) is applied to word embeddings to produce sentence embeddings, as shown in figure \ref{fig:methods_representations}.
% (\todo{I experienced w/ embedding based DTW as a similarity function, not sure how relevant it is to mention it, would need at least a ref})

By default, both $\T$ and $\R$ are sets of sentences\footnote{Sentences are determined by punctuation, which is predicted by the speech recognition system on the transcription side.}, however we also use sliding windows with overlap over sentences. For each document $D\in\left\{\T, \R \right\}$, the $k$-th sliding window $W_{o,s}^D$ is a set of $s$ sentences having its first (respectively last) $o$ sentences in common with previous window (resp. next). 
% a sliding window over of size $s$ and overlap $o$ contain $s$ sentences, and shares $o$ sentences with previous, and next windows.

% \begin{equation*}
% sliding\_window_{o, s}^{D} = \left\{ \sum_{i=k s - k o}^{(k+1)s-k o}{s_i} \mid k\in[1, \lvert D \rvert], s_i \in D\right\}
% \end{equation*}

% \begin{equation*}
% W_{o, s}^{D}(k) = \left\{s_i \mid i\in \left[k s - k o, (k+1)s-k o \right], s_i \in D\right\}
% \end{equation*}

\begin{equation}
W_{o, s}^{D}(k) = \left\{s_{ks - k o}, ..., s_{(k+1)s-k o} \mid s_i \in D\right\}
\label{eq:sliding_windows_def}
\end{equation}

Sliding windows aggregate sentence representations into a single vector using the $agg$ function (see figure \ref{fig:methods_representations}, then we calculate scores for all pairs of sliding windows from both sides:
\begin{equation}
S_{k, l}^{sliding} = \score\left(agg(W_{o, s}^{T}(k)), agg(W_{o, s}^{R}(l)) \right)
\label{eq:sliding_windows_agg}
\end{equation}

then similarities are assigned back to the sentence level:
\begin{equation}
    S_{i, j} = red\left(\left\{S_{k, l}^{sliding} \mid (s_i, s_j) \in \left(W_{o,s}^T(k)\times W_{o,s}^R(l) \right) \right\}\right)
\label{eq:sliding_windows_red}
\end{equation}
The reduction function $red$ (sum or product) calculates sentence scores from the sliding windows that contain it.

% \begin{equation*}
%     \score 
% \end{equation*}

% \subsection{Scoring}
% \notes{
%         \begin{enumerate}
%             \item Sentence based vs. SlidingWindow based
%             \item For texts: ROUGE
%             \item For tf-idf vectors: Latent Semantic Analysis (LSA) 
%             \item For embeddings: Cosine similarity
%             \item For embeddings: Dynamic Time Warping (DTW)
%         \end{enumerate}
% }    

\subsection{Alignment}
Having sentence level (of sentence-windows) similarities of every pairs of transcription and report, the alignment task is now to maximize the similarity at the document level.
We use dynamic programming, which aims to find an optimal path in the similarity matrix while ensuring -- by design -- that transcription and report are aligned chronologically.

We introduce the \textit{alignment matrix} $A$ that, for each coordinate $(i, j)$ corresponds to the similarity (eventually scaled to the power of $p$) plus the maximal value from its top $(i, j-1)$ or left $(i-1, j)$ neighbor coordinates :
\begin{align} 
\begin{split}
    A_{i, j} &= {S_{i, j}}^p + \max(A_{i-1, j}, A_{i, j-1})
\\  A_{1, 1} &= {S_{1, 1}}^p
% \\  \text{with }& p\in \left\{1;2;4\right\}
\end{split}
\label{eq:align_alignment}
\end{align}

At each position $(i, j)$ we keep track of the previous position (e.g. either $(i-1, j)$ or $(i, j-1)$): 
\begin{align} 
    H_{i, j} &= \argmax_{c \in \left\{(i-1, j); (i, j-1)\right\}}(A_c)
\label{eq:align_history}
\end{align}
ultimately giving us the optimal path, which correspond to the sentence level alignment $P$:
\begin{align*} 
    P_{k-1}  &= H_{i, j}\text{ with }(i, j) = P_k
\\  P_{\lenT + \lenR}  &= (\lenT, \lenR)
\end{align*}
% From this sentence-level alignment, we derive the segment-level alignment by considering a transcription segment $t_i$ to be aligned with the report segment $r_j$ that maximize mutual similarity alogn the path:
Figure \ref{fig:methods_alignment} shows simple example of the alignment process.

To derive the segment-level alignment of a transcription segment $\tseg_i$ we choose $\rseg_j$ to maximize similarity along the path:

\begin{align}
    \alignmentfct(m)=\argmax_{n} \left( \sum_{s_i \in \tseg_m}{\sum_{s_j \in \rseg_n} A_{i, j} \mathbbm{1}_{(i, j)\in P}} \right)
\label{eq:align_fct}
\end{align}

\begin{figure*}[h]  
% \subfloat[second caption here]
% {
\usetikzlibrary{matrix}
% \begin{minipage}{0.45\textwidth}
\centering
\begin{tikzpicture}[>=Computer Modern Rightarrow]
\matrix (A) [matrix of nodes,row sep=3mm,column sep=3mm,text height=0.2cm,anchor=center,
            nodes={align=center,text width=0.7cm},
        row 1/.style={anchor=center},
        column 1/.style={nodes={text width=0.5cm,align=right}}
]
{
  & $t_1$   & $t_2$     & $t_3$     & $t_4$    \\
$r_1$ & 5   & 3     & 8    & 9 \\
$r_2$ & 5   & 7     & 6     & 2 \\
$r_3$ & 3   & 4    & 7  & 5   \\
};
% \draw (A-4-1.south west)rectangle({A-2-5.north east}|-{A-1-2.north});
\draw (A-1-2.north west)--(A-4-2.south west);
\draw ([yshift=2mm]A-2-1.north west)--([yshift=2mm]A-2-5.north east);
\begin{scope}[thick,red,->]

\end{scope}
\end{tikzpicture}
% \end{minipage} % \hfill
% \begin{minipage}{0.45\textwidth}
\centering
\usetikzlibrary{matrix}
\begin{tikzpicture}[>=Computer Modern Rightarrow]

\matrix (A) [matrix of nodes,row sep=3mm,column sep=3mm,text height=0.2cm,anchor=center,
            nodes={align=center,text width=0.7cm},
        row 1/.style={anchor=center},
        column 1/.style={nodes={text width=0.5cm,align=right}}
    ]
{
  & $t_1$   & $t_2$     & $t_3$     & $t_4$    \\
$r_1$ & 5   & 8     & 16    & 25 \\
$r_2$ & 10   & 17     & 23     & 27 \\
$r_3$ & 13   & 21    & 30  & 35   \\
};
% \draw (A-4-1.south west)rectangle({A-2-5.north east}|-{A-1-2.north});
\draw (A-1-2.north west)--(A-4-2.south west);
\draw ([yshift=2mm]A-2-1.north west)--([yshift=2mm]A-2-5.north east);
\begin{scope}[thin,black,->]
% first line
\draw (A-2-2)--(A-2-3);
\draw (A-2-3)--(A-2-4);
\draw (A-2-4)--(A-2-5);

% first col
\draw (A-3-2)--(A-4-2);

% other suboptimal paths
\draw (A-2-5)--(A-3-5);
\draw (A-3-3)--(A-4-3);

\end{scope}
\begin{scope}[thick,red,->]
\draw (A-2-2)--(A-3-2);
\draw (A-3-2)--(A-3-3);
\draw (A-3-3)--(A-3-4);
\draw (A-3-4)--(A-4-4);
\draw (A-4-4)--(A-4-5);
\end{scope}
\end{tikzpicture}
% }
% \caption{Example Similarity Matrix $S$}
% \end{minipage}% \hfill

\captionsetup{width=0.8\textwidth}
\caption{Example of Dynamic Programming algorithm finding optimal path. At each coordinates $(i, j)$, the alignment ($A$, on the right) adds the corresponding similarity from $S$ (on the left) to highest neighbor value, either from top $(i, j-1)$ or left $(i-1, j)$, as shown with arrows (equivalent to $H$); red arrows represent the optimal path $P$. Similarity values are arbitrary here for simplicity.
\label{fig:methods_alignment}}
\end{figure*}

\subsection{Evaluation}
Linear segmentation performance is measured using \texttt{WindowDiff} \cite{Pevzner}, which compares boundaries predicted by the algorithm to the reference in a moving window of size $k$. \texttt{WindowDiff} is based on $P_k$ \cite{Beeferman1999} but is meant to be fairer, with respect to false negatives, number of boundaries, segment size and near miss errors. We report \texttt{WindowDiff} scores for our experiments.
We also consider simple metrics such as the segment accuracy and word accuracy. Experience scores are micro-averaged over reference files.

% \notes{
%     Metrics: 
%     \cite{Pevzner}: tl;dr: discuss metrics in segmentation tasks; other that precision, recall or so-called metric $P_k$ they introduce \texttt{WindowDiff} that does not penalize false-negative as hard (quite out topic, just that I considered WindowDiff at some point)
%     \begin{enumerate}
%         \item segment accuracy considered windowdiff as well (maybe pk also), without really different results.
%         \item word accuracy, word micro-accuracy
%         \item \todo{possibly filtered-word accuracy}
%     \end{enumerate}
% }

% \begin{figure}[!h]
%     \begin{center}
%         %\fbox{\parbox{6cm}{
%         %This is a figure with a caption.}}
%         \includegraphics[scale=0.5]{lrec2020W-image1.eps} 
%         \caption{The caption of the figure.}
%         \label{fig.1}
%     \end{center}
% \end{figure}

\section{Experiments}

\subsection{Bootstrapping the corpus creation}
    To build a corpus from scratch we iterate over three phases, (i) generating pre-alignments from the data using an automatic alignment model; (ii) correct the pre-alignment thanks to human annotators to get a gold reference set; (iii) evaluate models with respect to the new reference set. 
    
    Iterations increase the amount of gold references, allowing accurate evaluation of automatic alignment models, eventually making the annotators' task easier.
    
    % Models are evaluated before the correction -- which we call a priori score -- and after -- a posteriori score -- on the bigger reference set. The first pre-alignments does not have a priori scores since there was no references at the time.
    
% \notes{
%     GENERAL METHODOLOGY: working iteratively (aka. bootstrapping?!) 
%     \begin{enumerate}
%         \item produce alignements from a model (which maximize à priori score)
%         \item people fix alignments manually, providing gold-reference (and posteriori scores for the model)
%         \item evaluate models based on those references, goto 1
%     \end{enumerate}
% }

\paragraph{Gold Alignments} We developed an ad-hoc platform to collect gold alignments thanks to human annotators to serve as reference sets. We use our automatic alignment models to provide a pre-alignment that is then corrected by the annotator. 

\paragraph{Grid Search} In order to evaluate a wide variety of parameters at a reasonable computational cost,
we use several validation sets varying in their amount of reference files. The evaluation process iteratively selects best parameters, thus reducing their number, then evaluates these sub-sets on a bigger reference set. It helps us to efficiently explore the parameter space without spending too much effort on obviously sub-performing parameter sets and eventually identify most critical parameters.

% we first evaluate every parameters combinations against a single reference files, then only evaluate configurations that reach a given accuracy threshold on a dozen references, and so with bigger and bigger reference set. We eventually ends up fine-tuning with in-depth exploration of just a few parameters on the best performing configuration

\paragraph{1\textsuperscript{rst} Iteration: diagonal alignment} The first iteration started without any reference file, therefore, we had no way of quantitatively evaluating the auto-alignment process. Still, in order to provide a pre-alignment to human annotator, we used a naive approach that aligns segments diagonally: we do not compute similarity ( $S_{i,j} = 1, \forall (i, j)$) and move into the alignment matrix to stay on a diagonal
i.e. we replace the position history matrix $H$ of eq. \ref{eq:align_history} to be:
\begin{align*}
    H_{i, j}&=
        \left\{
            \begin{array}{ll}
                (i-1, j)\text{ if }r_{i, j} < r \\
                (i, j-1)\text{ otherwise }\\
            \end{array}
        \right.
% \\  \text{with }r &= \frac{\mid T \mid}{\mid R \mid}  \text{ and }r_{i,j} = \frac{i-1}{j-1}
\\  \text{with }r &= \mid T \mid / \mid R \mid 
\\ \text{ and }r_{i,j} &= (i-1) / (j-1)
% \label{eq:align_fct}
\end{align*}

\paragraph{2\textsuperscript{nd} Iteration: exploring scoring functions}
During the second iteration we mainly explored different sentence representations and scoring functions. Using plain text, we measure $ROUGE$ scores \cite{Lin2004}, more precisely R1-F, R2-F, and RL-F. We use vector representations of text based on (i) $tf\cdot idf$ and Latent Semantic Analysis; and (ii) pre-trained French word embeddings from \cite{fauconnier2015}, and score sentences based on cosine similarity. Word embeddings are trained with \texttt{word2vec} \cite{Mikolov2013c}. We experimented with both CBOW and Skip-Gram variants without significant performance differences.

Measuring similarities between sliding windows instead of sentences directly was meant to reduce impact of isolated sentences of low similarities. In fact, because our data don't perfectly match, there may be sentences with a very low similarity inside segments that actually discuss the same point. Parameters related to the sliding windows are the window size, and the overlap. We experimented with all combinations of $s \in \left\{0, 1, 2, 3, 4, 5, 10\right\}, o \in \left\{1, 2, 3, 5\right\}$. Related to its scores we consider \textit{aggregation} and \textit{reduction} function as parameters and experiment with $agg \in \left\{sum, mean, max\right\}$  and $red \in \left\{sum;product\right\}$.

\paragraph{3\textsuperscript{rd} Iteration: fine tuning embedding based models}
During the alignment phase, we found that the dynamic programming algorithm may keep the same direction for a long time. For example one report sentence may get high similarities with a lot of transcription sentences, resulting in a too monotonical alignment. To limit this behavior, we introduce horizontal and vertical decay factors (respectively $hd$ and $vd$), typically in $[0; 1]$, that lower scores in the same direction. We then consider a decayed alignment matrix $A'$ such as:
\begin{align}
\begin{split}
    A^{\prime}_{i, j}&= A_{i, j} \times D_{i, j}
\\  D_{i, j}&=
        \left\{
            \begin{array}{ll}
                D_{i-1, j} \times (1 - hd) \text{ if }A_{i-1,j} >  A_{i,j-1} \\
                D_{i, j-1} \times (1 - vd) \text{ otherwise } \\
            \end{array}
        \right.
% \\  \text{with }r &= \frac{\mid T \mid}{\mid R \mid}  \text{ and }r_{i,j} = \frac{i-1}{j-1}
% \\  \text{with }r &= \mid T \mid / \mid R \mid 
% \\ \text{ and }r_{i,j} &= (i-1) / (j-1)
\label{eq:align_decay}
\end{split}
\end{align}
The decay is reset to $D_{i, j} = 1$ at each change of direction.

% \todo{really not sure its clear, in dynamic programming, the score either comes from the top, or from the left as shown in eq. \ref{eq:align_alignment}, we use a vertical decay, horizontal decay and a moving decay. At each step decay = decay * hdecay or decay = decay * vdecay depending on last step. and score = score * decay}

\paragraph{4\textsuperscript{th} Iteration: \publicmeetings }
Finally, we select a set of public meetings in order to make it available for reproductions and benchmarks. This smaller corpus is used as a test set: no fine tuning has been done on this data for both the alignment and the summarization tasks. 

% We explore the parameter space with a kind of iterative grid search i.e. exploring a wide range of parameters with coarse graininess then exploring in more details when it make sense. In particular, lo
% \notes{
%     Kind of grid-search, not exhaustive tho (for obvious reasons), 
%     \begin{enumerate}
%         \item $o$, $s$, overlap and size of sliding window ($(0, 1)$ means no sliding-window), in eq.\ref{eq:sliding_windows_def}
%         \item $agg$, $red$, aggregation of sentence repr. into sliding windows, and reductions of scores, in eq.\ref{eq:sliding_windows_agg}, eq.\ref{eq:sliding_windows_red}
%         \item $p$ score power in alignment (whether to use the score, squared score, or score to the power of 4 during dynamic programming.
%         \item word embeddings to sentence aggregation: (not sure to mention, max == sum)
%     \end{enumerate}
% }

\subsection{Other models}
We also consider two linear segmentation baselines, namely \textsc{TextTiling} of \cite{Hearst1997} and \textsc{C99}\cite{Choi}. 

Linear segmentation baselines are penalized in comparison to our methods since they do not use the report document content. In particular, our methods cannot be wrong about the segment number since it is fixed by report side segmentation. Therefore, to make a fairer comparison, we only consider parameters sets that produce the excepted number of segments. Segment number can be explicitly set in \textsc{C99} whereas we had to grid search \textsc{TextTiling} parameters. 

\textsc{GraphSeg} from \cite{Glavas} has been considered, but producing long enough segments to be comparable with our work requires a low \textit{relatedness threshold}, which exponentially increases the computational cost.

% \todo{}
% \notes{
%     Models: 
%     \begin{enumerate}
%         \item TextTiling-best: we consider a simple baseline using the TextTiling algorithm. However, TextTiling is a mono-document segmentation algorithm therefore it does not make use of the similarity between the report and the transcription, nor is informed about how many segments are expected making the task harder by design. To compensate this a bit we run a large number of TextTiling segmentation exploring a large amount of parameters and keep the best score.
%         \item C99? 
%         \item GraphSeg? of \cite{Glavas}
%     \end{enumerate}
% }

\subsection{Summarization}
We trained neural summarization models on our data, first using gold set only, then incorporating automatically aligned data. Pre-processing include filtering segments based on their number of words and sentences, i.e. we consider segments if $10 \leq \#words \leq 1000$ and $3 \leq \#sentences \leq 50$.

Using \texttt{OpenNMT-py}\footnote{\url{https://github.com/OpenNMT/OpenNMT-py}}\cite{2017opennmt} we train \textit{Transformer} models \cite{Vaswani2017} similar to the baseline presented in \cite{ziegler2019} with the difference that we do not use any copy-mechanism.

Evaluation is conducted against our \texttt{public\_meetings} test set and uses the ROUGE-F metric \cite{Lin2004}.
 
\section{Results}

% \notes{
%     \begin{enumerate}
%         \item 1rst iteration: dummy model, scoring a posteriori on small refset around 10 docs
%         \item 2nd iteration: exploring some parameters, present priori/posteriori scores
%         \item 3rd iteration: idem
%         \item maybe present post-processing percent (i.e. segment changed/word changed)
%         \item maybe present à-priori/à-posteriori scores convergence which means the reference set is relevant
%         \item maybe present work about score/quality correlation i.e. validate models based on which gives the most reliable score i.e. score from which we could infer alignement quality, in order to filter.
%         \item Dataset stats, ubi-manual, ubi-auto, public-manual.
%     \end{enumerate}
% }

\subsection{Automatic Alignment Evaluation}
% TF-IDF & ROUGE are decieving while DIAGONAL makes a strong baseline
Table \ref{table:results_all} compares performances of automatic alignment models. 

Diagonal baseline shows interesting performances. In particular, it outperforms by a large margin linear segmentation algorithms and both of our tf-idf and ROUGE based models.

% Embedding approach largerly dominates
%   - sentence pooling has been explored, max and mean pooling, sum in marginaly better
%   - it seems that decay is a good idea, especially vertical decay, probably because our data is asymetrical in this axis therefore, discouraging verticality make sense
Embeddings based approaches are on a totally different level, with performances twice better than the diagonal baseline, and more than three times better than any other considered algorithm on the validation set.  

Introducing decays at the alignment stage is meant to avoid the alignment to be too monotonic. We started experimenting with small decays on both horizontal and vertical axes. Results make it clear that decays are key parameters. In particular, we found \textit{vertical decay} ($vd$) to have a greater impact, while \textit{horizontal decay} ($hd$) should be turned-off for maximal performances.
Similarly, scaling scores to the power of $p>1$ during alignment improves every model. In fact, it helps the model to distinguish good scores from average ones.

Sliding windows performs better that sentence representation (i.e. $s=1,o=0$) in most case -- only tf-idf models reach its top scores without it. However, we observed many different configurations of sizes, overlaps, aggregations and reduction functions reach high scores.

\begin{table*}[ht!]
\small
 \begin{center}
  \begin{tabular}{*{8}{c}}% c |}
    \toprule
        \subhead{Model}{\hfill}
        & \subhead{Window}{$(s, o)$}
        & \subhead{Window Scoring}{$(agg, red)$}
        & \subhead{Alignment}{$(hd, vd, p)$}
        &  \subhead{Dev. Acc. ($\mu$-avg) $\uparrow$}{$(seg. \%, word \%)$}
        %}
        & \subhead{Dev. WD $\downarrow$}{  \hfill  }
        & \subhead{Test Acc. ($\mu$-avg) $\uparrow$}{$(seg. \%, word \%)$}
        & \subhead{Test WD $\downarrow$}{  \hfill  }
\\ \midrule
        TextTiling
        & --
        & -- 
        & -- 
        & --
        & --
        % & 06.34 -- 05.37
        % & 37.06
        & 09.36 -- 06.66
        & 39.61 
\\  
        C99
        & --
        & -- 
        & -- 
        & --
        & --
        % & 06.34 -- 05.37
        % & 37.06
        & 05.68 -- 05.03
        & 42.49       
% \\  
%         GraphSeg
%         & --
%         & -- 
%         & -- 
%         & seg -- word
%         & wd
\\ \midrule
        Diagonal
        & --
        & -- 
        & -- 
% FULL: 20.232 	18.590 	21.555 	22.723 	
% PUB: 27.243 	20.748 	23.287 	24.662 	34.619
        & 18.59 -- 21.55
        & 17.43
        & 20.75 -- 23.28
        & 34.61
\\
\submidrule

        \multirow{3}{*}{tf-idf}
        % 2 	1 	sum 	sum 	pow4 	1.0 	1.0 	2.019 	1.907 	2.029 	2.238 	35.698
            & 2 -- 1 
            & sum -- sum 
            & $ 0 $ ; $ 0 $ ; 4
            & 5.02 -- 5.23
            & 13.80
            & 4.10 -- 5.17
            & 23.02

    \\
% FULL: sum 	10 	3 	mean 	mul 	identity 	n/a 	n/a 	10.449 	9.694 	10.909 	11.500 	17.326
% PUB : sum 	10 	3 	mean 	mul 	identity 	n/a 	n/a 	12.622 	9.910 	10.946 	11.592 	33.004
            & 10 -- 3
            & mean -- mul
            & $10^{-4}$ ; $10^{-4}$ ; 1
            & 9.69 --	10.90
            & 17.33
            & 9.91 -- 10.94  	
            & 33.00
    \\
% FULL: 12.458 	10.928 	12.326 	12.994 	17.744
% PUB: 9.458 	10.289 	10.896 	35.047
            & 1 -- 0
            & max -- prod 
            & $10^{-4}$ ; $10^{-4}$ ; 2
            & 10.93 -- 12.33
            & 17.74
            & 10.29 -- 10.90
            & 35.05
\\  
\submidrule
        \multirow{2}{*}{ROUGE}
% FULL: 11.430 	12.701 	13.389 	17.157
% PUB : 9.252 	9.664 	10.235 	32.994
            & 10 -- 2
            & cat -- sum 
            & $10^{-4}$ ; $10^{-4}$ ; 1
            & 11.43 -- 12.70
            & 17.157
            & 9.25 -- 9.66
            & 32.99
    \\
% FULL: 14.519 	16.954 	17.873 	17.847
% PUB : 10.632 	11.766 	12.461 	34.016
            & 2 -- 0
            & cat -- sum 
            & $10^{-4}$ ; $10^{-4}$ ; 4
            & 14.52 -- 16.95
            & 17.85
            & 10.63 -- 11.76
            & 34.01 
    \\
\submidrule
        \multirow{3}{*}{Embeddings}
% FULL: 45.622 	54.064 	56.993 	10.813
% PUB:  64.361 	72.950 	77.257 	19.872

            % #28(26) 	sum 	2 	1 	sum 	mul 	pow4 	1 	1 	38.738 	39.754 	48.880 	53.911 	32.05
            & 2 -- 1
            & sum -- prod
            & $ 0 $ ; $ 0 $ ; 4 	
            & 45.62 -- 54.06
            & 10.81
            & 64.36 --72.95
            & 19.872 
    \\
% FULL: 60.964 	70.729 	74.562 	10.941
% PUB : 58.968 	68.087 	72.108  23.432
            & 2 -- 1
            & sum -- prod 
            & $10^{-4}$ ; $10^{-4}$ ; 4
            & 60.96 -- 70.73
            & 10.94
            & 58.97 -- 68.09
            & 23.43
    \\    
% FULL: 60.992 	72.490 	76.418 	10.380
% PUB : 69.355 	79.060 	83.728 	15.089
            & 2 -- 1
            & sum -- prod 
            & $ 0 $ ; $10^{-4}$ ; 4
            & \textbf{61.00} -- \textbf{72.50}
            & \textbf{10.38}
            & \textbf{69.36} -- \textbf{79.06}
            & \textbf{15.09}

\\ \bottomrule
  \end{tabular}
  \caption{Automatic alignment models evaluation against the validation set (202 reference meetings) and \publicmeetings~test set (22 meetings) on three metrics: segment accuracy, word accuracy, and WindowDiff. \label{table:results_all}}
 \end{center}
 
\end{table*}

\subsection{Human Evaluation}
Human annotators align transcription segments with respect to report segments based on a pre-alignment produced by automatic alignment models. 
As we were fine tuning our models, we provided better pre-alignments, eventually making the annotator's task easier The alignment process for the annotators consists in checking the pre-alignment and correcting mismatches one segment at a time. We report human evaluated segment-level accuracy as the ratio of segments that were not modified by the annotator against the total number of segments.

\begin{figure}[h]
\begin{center}
%\fbox{\parbox{6cm}{
%This is a figure with a caption.}}
% \includegraphics[scale=0.5]{fig/postedit_similarity.png} 
\includegraphics[width=\linewidth]{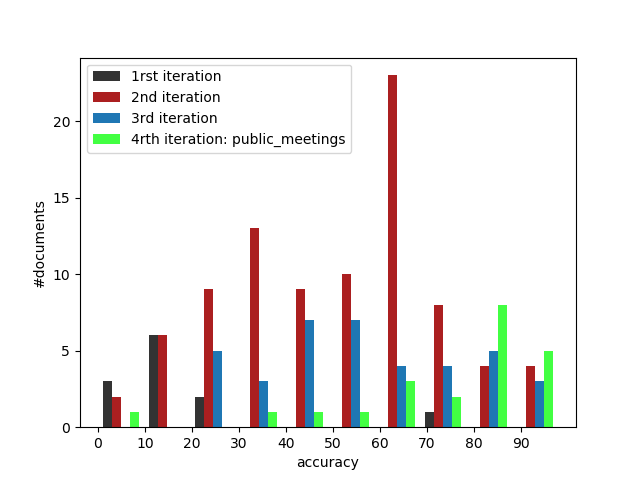} 

\caption{Annotator evaluation with respect to the automatic pre-alignment for each iterations}
\label{fig:results_postedit_similarity}
\end{center}
\end{figure}

Figure \ref{fig:results_postedit_similarity} and table \ref{table:results_postedit_similarity} show, for each iteration, the accuracy distribution. We observe that accuracy is consistently increasing over iterations.

\begin{table}[H]
\begin{center}
\begin{tabular}{*{4}{c}}
\toprule
    & \subhead{\#documents}{\hfill} & \subhead{Annotator Score $\uparrow$}{$(mean, median)$}

\\\midrule    1rst iteration 
                & 12 & 18.63 -- 15.73     
\\    2rst iteration 
                & 88 & 50.44 -- 53.56  
\\    3rd iteration 
                & 38 & 57.23 -- 55.02  
\\    \publicmeetings
                & 22 & 72.67 -- 80.08  
\\\bottomrule
\end{tabular}
\caption{Human evaluation of automatic alignments}
\label{table:results_postedit_similarity}
 \end{center}
\end{table}

\subsection{Summarization}
Summarization models have first been trained on human annotated alignments only, then with a larger dataset that also contains $70,000$ more training pairs emanating from automatic alignment. We find that using automatic alignment for data annotation makes a substantial difference in the summarization performance of almost +4 ROUGE points  (table \ref{table:results_summarization_publicmeetings}).
This result is encouraging and motivates us to continue automatic annotation.

\begin{table}[H]
\begin{center}
\begin{tabular}{lcc} % c|}
\toprule
    \subhead{Dataset}{\hfill}
        & \subhead{\#Pairs}{$(train, test)$}
        % & \subhead{Training Lengths}{$(src, tgt)$}
        & \subhead{ROUGE Score (F)}{$(R1, R2, RL)$}
\\\midrule    Gold dataset 
                & 21k -- 1060
                % & ... -- ... 
                & 52.80 / 29.59 / 49.49
\\          Gold + Automatic 
                & 91k -- 1060 
                & 56.56 / 35.43 / 53.55

\\\bottomrule
\end{tabular}
\caption{Scores on the \publicmeetings  test set of automatic summarization models trained on human references only vs. extend the dataset with annotations from automatic alignment}
\label{table:results_summarization_publicmeetings}
 \end{center}
\end{table}

% \begin{table*}[ht]
% \begin{center}
% \begin{tabular}{|l|l|}

% \end{tabular}
% \caption{The caption of the big table}
% \end{center}
% \end{table*}

% \subsection{Tables}

% The instructions for tables are the same as for figures.
%Two types of tables are distinguished: in-column and big tables that don't fit in the columns.
%\subsection{In-column tables}
%An example of an in-column table is presented here.
%
% \begin{table}[!h]
% \begin{center}
% \begin{tabularx}{\columnwidth}{|l|X|}

%       \hline
%       Level&Tools\\
%       \hline
%       Morphology & Pitrat Analyser\\
%       \hline
%       Syntax & LFG Analyser (C-Structure)\\
%       \hline
%      Semantics & LFG F-Structures + Sowa's\\
%      & Conceptual Graphs\\
%       \hline

% \end{tabularx}
% \caption{The caption of the table}
%  \end{center}
% \end{table}

% \subsection{Big tables}

% An example of a big table which extends beyond the column and will
% float in the next page.

% \begin{table*}[ht]
% \begin{center}
% \begin{tabular}{|l|l|}

%       \hline
%       Level&Tools\\
%       \hline\hline
%       Morphology & Pitrat Analyser\\
%       Syntax & LFG Analyser (C-Structure)\\
%       Semantics & LFG F-Structures + Sowa's Conceptual Graphs  \\
%       \hline

% \end{tabular}
% \caption{The caption of the big table}
% \end{center}
% \end{table*}

\section{Discussion and Future Work}

% To get the highest quality alignments from human annotators, we asked them to only align transcription segments that are actually reported. In fact, recordings may contain interventions that are not present in the report because they are judged irrelevant by the writer. 

During the alignment process, we make the assumption that each transcription segment must be aligned. However, in practice we asked human annotators to filter out irrelevant segments. Such segments are part of the validation set, but flagged in order that they should not be assigned to any report segments. During evaluation we penalize models for each false alignment assigned to irrelevant segments so that our results are comparable to future models capable of ignoring some transcription segments. To get an idea of how important this phenomenon is, we adapt word accuracy to ignore irrelevant segments and find a $4.7\%$ absolute difference.

\begin{align*}
    word\_acc &= \frac{\#W_{aligned}}{\#W}
\\  pos\_word\_acc &= \frac{\#W_{aligned}}{\#W - \#W_{irrelevant}}
\end{align*}

Word embedding vectors used in this work have been trained by \cite{fauconnier2015} who made them publicly available. \footnote{More information and vectors can be found at \url{https://fauconnier.github.io/\#data}}. While they make our results fully reproducible, training embedding vectors on our data would be an interesting area for future research and could improve the quality of the automatic alignment. 

Lastly, we would like to study whether the alignment scores provided by our models could be used to predict the alignment quality. Such predictions could be used to filter automatic annotations and use only the potentially relevant automatically aligned segments.

% \notes{
%     \begin{enumerate}
%         \item ABOUT negative segments, we find pos accuracy += x prct on embeddings models. Our models does not eliminate bad segments
%         \item current summarization is document independant. In particular it may be improved with context, as simple as a named entity list for eg.
%         \item WD is weird
%         \item can be computationally way cheaper (we calculate a whole $m \times n$ similarity matrix but in practice only use the diagonal (as shown by diagonal baselines)
%         \item FUTURE WORK: better word embedding could help, but using public ones helps reproducibility/future comparison, \todo{We could do it now, I already have LM trained on ubiqus data, would give an interesting comparison/advice whether using specific embeddings leads to improvement}
%         \item FUTURE WORK: eventually explore neural approaches, now that we get some data (and a way to augment it)
%         \item FUTURE WORK: other data-augmentation: back-sum
%         \item Explore alignment of other
%         \item 
%         \item 
%     \end{enumerate}
% }

\section{Conclusion}

This paper has explored the development of automatic alignment models to map speaker interventions from meeting reports to corresponding sentences of a transcription. Meetings last several hours, making them unsuitable sources for training as-is; therefore, segmentation is a key pre-processing step in neural approaches for automatic summarization.
Our models align transcription sentences -- as provided by our speech recognition system -- with respect to report segments, delimited by tags in the document (either in the header or when explicitly specifying a change of speaker).

We introduce \publicmeetings~, a novel meeting summarization corpus against which we evaluate both automatic alignment and summarization.

We have shown that our automatic alignment models allow us to greatly increase our corpus size, leading to better summarization performance on all $ROUGE$ metrics (R1, R2, RL).

% Your submission of a finalised contribution for inclusion in the LREC
% proceedings automatically assigns the above-mentioned copyright to ELRA.

% \section{Acknowledgements}

% Place all acknowledgements (including those concerning research grants and
% funding) in a separate section at the end of the article.

% \section{Providing References}

% \subsection{Bibliographical References}
% Bibliographical references should be listed in alphabetical order at the
% end of the article. The title of the section, ``Bibliographical References'',
% should be a level 1 heading. The first line of each bibliographical reference
% should be justified to the left of the column, and the rest of the entry should
% be indented by 0.35 cm.

% The examples provided in Section \secref{main:ref} (some of which are fictitious
% references) illustrate the basic format required for articles in conference
% proceedings, books, journal articles, PhD theses, and chapters of books.

% \subsection{Language Resource References}

% Language resource references should be listed in alphabetical order at the end
% of the article.

% \section*{Appendix: How to Produce the \texttt{.pdf} Version}

% In order to generate a PDF file out of the LaTeX file herein, when citing
% language resources, the following steps need to be performed:

% \begin{itemize}
%     \item{Compile the \texttt{.tex} file once}
%     \item{Invoke \texttt{bibtex} on the eponymous \texttt{.aux} file}
%  %   \item{Invoke \texttt{bibtex} on the \texttt{languageresources.aux} file}
%     \item{Compile the \texttt{.tex} file twice}
% \end{itemize}

% % \nocite{*}

\section{Bibliographical References}
\label{main:ref}

\bibliographystyle{lrec}
\bibliography{bib/TextAlignment, bib/Summarization2019, bib/misc_bib}

%\section{Language Resource References}
%\label{lr:ref}
%\bibliographystylelanguageresource{lrec}
%\bibliographylanguageresource{lrec2020W-xample}

\end{document}